\documentclass[11pt,a4paper]{article}
\usepackage[hyperref]{emnlp-ijcnlp-2019}
\usepackage{times}
\usepackage{latexsym}
\usepackage{tikz}
\usepackage{mathtools}
\usepackage{url}
\usepackage{multirow}

\aclfinalcopy 

\makeatletter
\newcommand{\printfnsymbol}[1]{%
  \textsuperscript{\@fnsymbol{#1}}%
}
\makeatother
\title{Efficient Sentence Embedding using Discrete Cosine Transform}

\author{Nada Almarwani,\thanks{\ \ Both authors contributed equally.} $^{\textbf{1,2}}$ Hanan Aldarmaki,\printfnsymbol{1}$^\textbf{3}$ Mona Diab $^{\textbf{1,4}}$\\
$^1$ Dept. of Computer Science, The George Washington University\\
$^2$ Dept. of Computer Science, Taibah University\\
$^3$ Computer Science \& Software Engineering Dept., UAEU\\
$^4$ Amazon AWS AI\\
{\tt nadaoh@gwu.edu, h-aldarmaki@uaeu.ac.ae, diabmona@amazon.com}
}

\date{}

\begin{document}
\maketitle
\begin{abstract}
Vector averaging remains one of the most popular sentence embedding methods in spite of its obvious disregard for syntactic structure. While more complex sequential or convolutional networks potentially yield superior classification performance, the improvements in classification accuracy are typically mediocre compared to  simple vector averaging. 
As an efficient alternative, we propose the use of discrete cosine transform (DCT) to compress word sequences in an order-preserving manner. The lower order DCT coefficients represent the overall feature patterns in sentences, which results in suitable embeddings for tasks that could benefit from syntactic features. Our results in semantic probing tasks demonstrate that DCT embeddings indeed preserve more syntactic information compared with vector averaging. With practically equivalent complexity, the model yields better overall performance in downstream classification tasks that correlate with syntactic features, thereby illustrating the capacity of DCT to preserve word order information. 
\end{abstract}

\section{Introduction}
Modern NLP systems rely on word embeddings as input units to encode the statistical semantic and syntactic properties of words, ranging from standard context-independent embeddings such as word2vec \cite{mikolov2013distributed} and Glove \cite{pennington2014glove} to contextualized embeddings such as ELMo \cite{peters2018deep} and BERT \cite{devlin2018bert}. However, most applications operate at the phrase or sentence level. Hence, the word embeddings are averaged to yield sentence embeddings. Averaging is an efficient compositional operation that leads to good performance. In fact, averaging is  difficult to beat by more complex compositional models as illustrated across several classification tasks: topic categorization, semantic textual similarity, and sentiment classification \cite{aldarmaki2018evaluation}. Encoding sentences into fixed-length vectors that capture various full sentence linguistic properties leading to performance gains across different classification tasks remains a challenge. Given the complexity of most models that attempt to encode sentence structure, such as convolutional, recursive, or recurrent networks, the trade-off between efficiency and performance tips the balance in favor of simpler models like vector averaging.
Sequential neural sentence encoders, like Skip-thought \cite{kiros2015skip} and InferSent \cite{conneau-EtAl:2017:EMNLP2017}, can potentially encode rich semantic and syntactic features from sentence structures. However, for practical applications, sequential models are rather cumbersome and inefficient, and the gains in performance are typically mediocre compared with vector averaging \cite{aldarmaki2018evaluation}. In addition, the more complex models typically don't generalize well to out-of-domain data \cite{wieting2016iclr}.
FastSent \cite{hill2016learning} is an unsupervised alternative approach of lower computational cost, but similar to vector averaging, it disregards word order. Tensor-based composition can effectively capture word order, but current approaches rely on restricted grammatical constructs, such as transitive phrases, and cannot be easily extended to variable-length sequences of arbitrary structures \cite{milajevs10evaluating}. Therefore, despite its obvious disregard for structural properties, the efficiency and reasonable performance of vector averaging makes them more suitable for practical text classification. 

In this work, we propose to use the Discrete Cosine Transform (DCT) as a simple and efficient way to model word order and structure in sentences while maintaining practical efficiency. DCT is a widely-used technique in digital signal processing applications such as image compression \cite{watson1994image}  as well as speech recognition \cite{huang2000dct}.
We use DCT to summarize the general feature patterns in word sequences and compress them into fixed-length vectors. Experiments in probing tasks demonstrate that our DCT embeddings preserve more syntactic and semantic features compared with vector averaging. Furthermore, the results indicate that DCT performance in downstream applications  is correlated with these features.

\section{Approach} 
\subsection{Discrete Cosine Transform}

Discrete Cosine Transform (DCT) is an invertible function that maps an input sequence of $N$ real numbers to the coefficients of $N$ orthogonal cosine basis functions. Given a vector of real numbers $\vec{v}={v_0,...,v_{N-1}}$, we calculate a sequence of DCT coefficients $c[0],...,c[N-1]$ as follows:\footnote{There are several variants of DCT. We use DCT type II multiplied by an overall scale factor, which makes the resulting DCT matrix orthogonal\cite{shao2008type} in our implementation}
\begin{equation}
 c[0]= \sqrt{\frac{1}{N}}\sum_{n=0}^{N-1}v_n,
\end{equation}
and 
\begin{equation}\label{eq:dct}
 c[k]=\sqrt{\frac{2}{N}}\sum_{n=0}^{N-1}v_n \cos{\frac{\pi}{N}(n+\frac{1}{2})k},
\end{equation}
for $1 \le k<N$. Note that $c[0]$ is the sum of the input sequence normalized by the square length, which is proportional to the average of the sequence. The $N$ coefficients can be used to reconstruct the original sequence exactly using the inverse transform. In practice, DCT is used for compression by preserving only the coefficients with large magnitudes. Lower-order coefficients represent lower signal frequencies which correspond to the overall patterns in the sequence \cite{ahmed1974discrete}.

\subsection{DCT Sentence Embeddings}
\label{detailed}
We apply DCT on the word vectors along the length of the sentence. Given a sentence of $N$ words ${w_1,...,w_N}$, we stack the sequence of $d$-dimensional word embeddings in an $N \times d$ matrix, then apply DCT along the column. In other words, each feature in the vector space is compressed independently, and the resultant DCT embeddings summarize the feature patterns along the word sequence. To get a fixed-length and consistent sentence vector, we extract and concatenate the first $K$ DCT coefficients and discard higher-order coefficients, which results in sentence vectors of size $Kd$. For cases where $N < K$, we pad the sentence with $K-N$ zero vectors.\footnote{Evaluation script is available at https://github.com/N-Almarwani/DCT\_Sentence\_Embedding} An illustration of the DCT embeddings process is given in Figure \ref{DCT-illustration}. In image compression, the magnitude of the coefficients tends to decrease with increasing $k$, but we didn't observe this trend in text data except that $c[0]$ tends to have larger absolute value than the remaining coefficients. Nonetheless, by retaining lower-order coefficients we get a consistent representation of overall feature patterns in the word sequence.\\
\begin{figure*}
\centering
\hspace{-10pt}
\includegraphics[width=\textwidth]{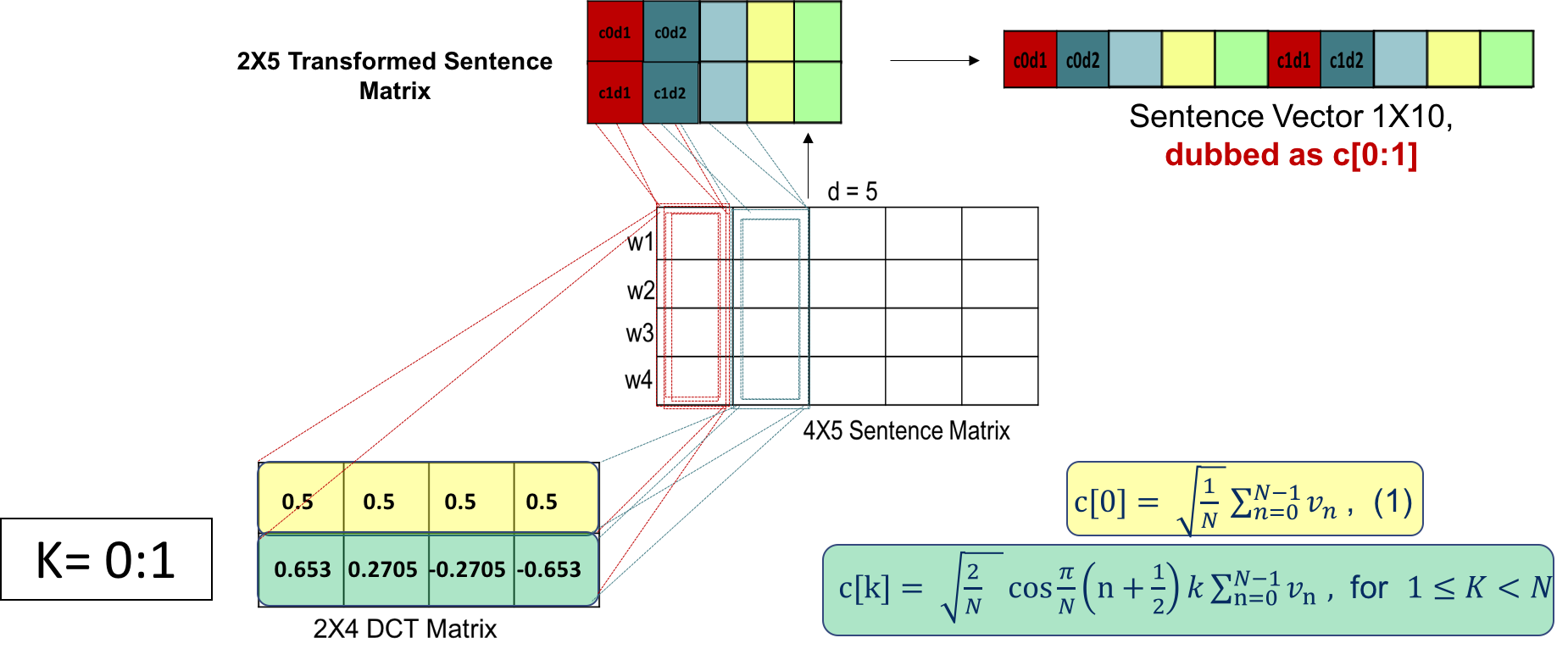}
\caption{An illustration of DCT embeddings. The size of the sentence to encode is \(4\times5\), where 4 is the number of words and 5 is the number of word embedding dimensions. Each feature vector is transformed using DCT independently. In this example, K=0:1 and the final representation is the concatenation of the first two coefficient from all transformed features.}
\label{DCT-illustration}
\end{figure*}
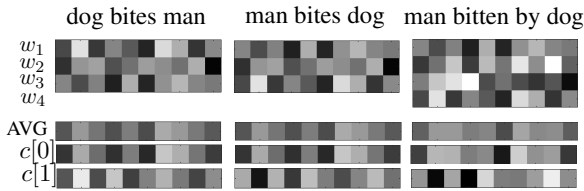
\begin{figure}
\hspace{-10pt}
\scalebox{0.85}{
\begin{tikzpicture}[x=0.75pt,y=0.75pt,yscale=-1,xscale=1]

\draw (370.98,136.75) node {\includegraphics[width=72.91pt,height=22.88pt]{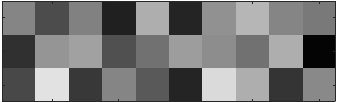}};
\draw (371.91,174.92) node {\includegraphics[width=73.41pt,height=8.13pt]{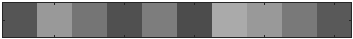}};
\draw (371.75,188.5) node {\includegraphics[width=73.63pt,height=8.5pt]{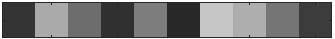}};
\draw (371.75,203) node {\includegraphics[width=73.4pt,height=8.7pt]{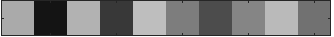}};
\draw (265.98,136.5) node {\includegraphics[width=73.59pt,height=23.25pt]{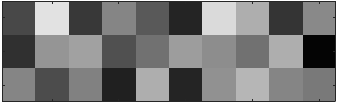}};
\draw (266.31,174.92) node {\includegraphics[width=73.41pt,height=8.13pt]{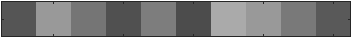}};
\draw (266.75,188.5) node {\includegraphics[width=73.63pt,height=8.5pt]{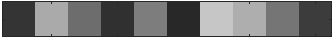}};
\draw (267,203) node {\includegraphics[width=73.4pt,height=8.7pt]{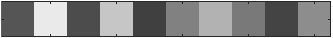}};
\draw (475.93,141.15) node {\includegraphics[width=73.21pt,height=30.22pt]{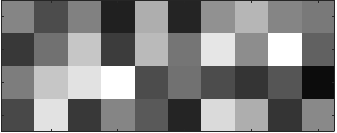}};
\draw (475.31,174.92) node {\includegraphics[width=73.41pt,height=8.13pt]{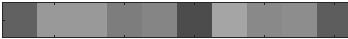}};
\draw (474.75,188.5) node {\includegraphics[width=73.63pt,height=8.5pt]{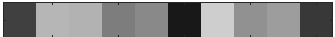}};
\draw (475,203) node {\includegraphics[width=73.4pt,height=8.7pt]{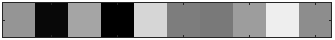}};

\draw (200.83,173.5) node [align=left] {{\small AVG}};
\draw (207.33,187.17) node [align=left] {{$c[0]$}};
\draw (206.67,200.83) node [align=left] {{ $c[1]$}};
\draw (370,107.67) node [align=left] {man bites dog};
\draw (265,108.33) node [align=left] {dog bites man};
\draw (475,108) node [align=left] {man bitten by dog};
\draw (204,126) node [scale=0.9] [align=left] {$w_1$};
\draw (204,136) node [scale=0.9] [align=left] {$w_2$};
\draw (204,145.5) node [scale=0.9] [align=left] {$w_3$};
\draw (204.5,156.5) node [scale=0.9] [align=left] {$w_4$};

\end{tikzpicture}
}
\caption{Illustration of word vector averaging vs. DCT using the first 2 DCT coefficients. The word vectors are generated randomly from a standard normal distribution with $d=10$.}\label{fig:demo}
\end{figure}

Figure \ref{fig:demo} illustrates the properties of DCT embeddings compared to vector averaging (AVG). Notice that the first DCT coefficients, $c[0]$, result in vectors that are independent of word order since the lowest frequency represents the average energy in the sequence. In this sense, $c[0]$ is similar to AVG, where ``dog bites man" and ``man bites dog" have identical embeddings. The second-order coefficients, on the other hand, are sensitive to word order, which results in different representations for the above sentence pair. The counterexample ``man bitten by dog" shows that $c[1]$ embeddings are most sensitive to the overall patterns---in this case: ``man ... dog"---which results in an embedding more similar to ``man bites dog", than the semantically similar ``dog bites man". However, there are still some variations in the final embeddings from the different word components (`bitten' vs. `bite'), which can potentially be useful in downstream tasks. Since both DCT and AVG are unparameterized, the downstream classifiers can incorporate a hidden layer to learn these subtle variations in higher-order features depending on the learning objective. 

\subsection{A Note on Complexity}
The cosine terms in Equation \ref{eq:dct} can be pre-calculated for efficiency. For a maximum sentence length $\hat{N}$ and a given $K$, the total number of terms is $(K-1)\hat{N}$ for each feature. The run-time complexity is equivalent to calculating $K$ weighted averages, which is proportional to $KN$, where $K$ should be set to a small constant relative to the expected length.\footnote{We experimented with $1 \le K \le 7$.} Note also that the number of input parameters in downstream classification models will increase linearly with $K$. With parallel implementations however, the difference in run-time complexity between AVG and DCT is practically negligible.

\section{Experiments and Results}

\subsection{Evaluation Framework}
We use the SentEval toolkit\footnote{https://github.com/facebookresearch/SentEval} \cite{conneau2018senteval} to evaluate the sentence representations on different probing as well as downstream classification tasks. The probing benchmark set was designed to analyze the quality of sentence embeddings. It contains a set of 10 classification tasks, summarized in Table \ref{Summary}, that address varieties of linguistic properties including surface, syntactic, and semantic information \cite{conneau2018you}. The downstream set, on the other hand, includes the following standard classification tasks: binary and fine-grained sentiment classification (MR, SST2, SST5) \cite{pang2004sentimental, socher2013recursive}, product reviews (CR) \cite{hu2004mining}, opinion polarity (MPQA) \cite{wiebe2005annotating}, question type classification (TREC) \cite{voorhees2000building}, natural language inference (SICK-E) \cite{marelli2014semeval}, semantic relatedness (SICK-R, STSB) \cite{marelli2014semeval, cer2017semeval}, paraphrase detection (MRPC) \cite{dolan2004unsupervised}, and subjectivity/objectivity (SUBJ) \cite{pang2004sentimental}. 

\begin{table}[h]
\centering
\setlength\tabcolsep{4pt}
\begin{tabular}{|l|l|}
\hline
\textbf{Task}&\textbf{Description}\\\hline
SentLen&Length prediction\\\hline
WC&Word Content analysis\\\hline
TreeDepth&Tree depth prediction\\\hline
TopConst&Top Constituents prediction\\\hline
BShift&Word order analysis\\\hline
Tense&Verb tense prediction\\\hline
SubjNum&Subject number prediction\\\hline
ObjNum&Object number prediction\\\hline
SOMO&Semantic odd man out\\\hline
CoordInv&Coordination Inversion\\\hline
\end{tabular}
\caption{Probing Tasks}
\label{Summary}
\end{table}

\begin{table*}[t]
\centering
\setlength\tabcolsep{2pt}
\begin{tabular}{c|cc|ccc|ccccc|}
\cline{2-11}
&\multicolumn{2}{c|}{\textbf{Surface}}&\multicolumn{3}{c|}{\textbf{Syntactic}}&\multicolumn{5}{c|}{\textbf{Semantic}}\\
\cline{2-11}
\hline
\multicolumn{1}{|c|}{Model}&SentLen&WC&TreeDepth&TopConst&BShift&Tense&SubjNum&ObjNum&SOMO&CoordInv\\\hline
\multicolumn{1}{|c|}{Majority}&20.0&0.5&17.9&5.0&50.0&50.0&50.0&50.0&50.0&50.0\\\hline
\multicolumn{1}{|c|}{Human}&100&100&84.0&84.0&98.0&85.0&88.0&86.5&81.2&85.0\\\hline
\multicolumn{1}{|c|}{Length}&100&0.2&18.1&9.3&50.6&56.5&50.3&50.1&50.2&50.0\\\hline
\hline
\multicolumn{1}{|c|}{AVG}&64.12&82.1&36.38&68.04&50.16&87.9&80.89&80.24&50.39&51.95\\\hline
\multicolumn{1}{|c|}{MAX}&62.67&88.97&33.02&62.63&50.31&85.66&77.11&76.04&51.86&52.33\\\hline
\multicolumn{1}{|c|}{$c[0]$}&\textbf{98.67}&\textbf{91.11}&38.6&70.54&50.42&88.25&80.88&80.56&\textbf{55.6}&55\\\hline
\multicolumn{1}{|c|}{$c[0:1]$}&97.18&89.16&40.41&78.34&52.25&88.58&86.59&84.36&54.62&70.42\\\hline
\multicolumn{1}{|c|}{$c[0:2]$}&95.84&86.77&43.01&80.41&54.84&88.87&88.06&86.26&53.07&71.87\\\hline
\multicolumn{1}{|c|}{$c[0:3]$}&94.63&84.96&\textbf{43.35}&81.01&57.29&88.88&88.36&86.51&53.79&\textbf{72.01}\\\hline
\multicolumn{1}{|c|}{$c[0:4]$}&93.25&83.24&43.26&81.49&60.31&\textbf{88.91}&\textbf{88.65}&87.15&52.77&71.91\\\hline
\multicolumn{1}{|c|}{$c[0:5]$}&92.29&81.84&42.75&\textbf{81.60}&62.01&88.82&88.44&87.98&52.38&70.96\\\hline
\multicolumn{1}{|c|}{$c[0:6]$}&91.56&79.83&43.05&81.41&\textbf{62.59}&88.87&\textbf{88.65}&\textbf{88.28}&52.07&70.63\\\hline
\end{tabular}
\caption{Probing tasks performance of vector averaging (AVG) and max pooling (MAX) vs. DCT embeddings with various $K$. Majority (baseline), Human (human-bound), and a linear classifier with sentence length as sole feature (Length) as reported in \cite{conneau2018you}, respectively.}

\label{ProbingBest}
\end{table*}


\begin{table*}[t]
\centering
\setlength\tabcolsep{4pt}
\begin{tabular}{c|ccccc|c|ccc|c|c|}
\cline{2-12}
& \multicolumn{5}{c|}{\textbf{Sentiment Analysis}}    & \multirow{2}{*}{\textbf{SUBJ}} & \multicolumn{3}{c|}{\textbf{Relatedness/Paraphrase}} & \textbf{Inference} & \multirow{2}{*}{\textbf{TREC}} \\ \cline{1-6} \cline{8-11}
\multicolumn{1}{|c|}{Model}& MR& SST2  & SST5  & CR & MPQA&    & SICK-R & STSB &MRPC&SICK-E &    \\ \hline
\multicolumn{1}{|c|}{AVG}&78.3&\textbf{84.13}&44.16&79.6&87.94&92.33&81.95&69.26&74.43&79.5&83.2\\ \hline
\multicolumn{1}{|c|}{MAX}&73.31&79.24&41.86&73.35&86.54&88.02&81.93&\textbf{71.57}&72.5&77.98&76.2\\ \hline
\multicolumn{1}{|c|}{$c[0]$}&\textbf{78.45}&83.53&44.57&79.81&\textbf{88.36}&\textbf{92.79}&82.61&71.11&72.93&78.91&84.8\\ \hline
\multicolumn{1}{|c|}{$c[0:1]$}&78.15&83.47&\textbf{46.06}&79.84&87.76&92.61&82.73&70.82&72.81&79.64&88.2\\ \hline
\multicolumn{1}{|c|}{$c[0:2]$}&78.02&82.98&45.16&79.68&87.62&92.5&\textbf{82.95}&70.36&72.87&79.76&\textbf{89.8}\\ \hline
\multicolumn{1}{|c|}{$c[0:3]$}&77.81&83.8&45.79&79.66&87.54&92.4&82.93&69.79&73.57&\textbf{80.56}&88.2\\ \hline
\multicolumn{1}{|c|}{$c[0:4]$}&77.72&83.75&44.03&\textbf{80.08}&87.4&92.61&82.53&69.31&72.35&79.72&\textbf{89.8}\\ \hline
\multicolumn{1}{|c|}{$c[0:5]$}&77.42&82.43&43.3&78.6&87.21&92.19&82.36&68.9&73.91&79.89&88.8\\ \hline
\multicolumn{1}{|c|}{$c[0:6]$}&77.47&82.81&42.99&78.78&87.06&92.15&81.86&68.17&\textbf{75.07}&79.76&86.4\\ \hline

\end{tabular}
\caption{DCT embedding Performance in SentEval downstream tasks compared to vector averaging (AVG) and max pooling (MAX). } 
\label{downstream}
\end{table*}

\begin{table*}
\centering
\begin{tabular}{c| c| c| c|c|c|c|c|c|c|c|}
\cline{2-10}
&\multicolumn{3}{|c|}{\textbf{20-NG}}&\multicolumn{3}{|c|}{\textbf{R-8}}&\multicolumn{3}{|c|}{\textbf{SST-5}}\\

\cline{3-10}
\hline
\multicolumn{1}{|c|}{Model}&P&R&F1&P&R&F1&P&R&F1\\\hline
\multicolumn{1}{|c|}{PCA}&55.43&54.67&54.77&83.83&83.42&83.41&26.47&25.08&25.23\\\hline
\multicolumn{1}{|c|}{DCT*}&61.07&59.16&59.78&90.41&90.78&90.38&30.11&30.09&29.53\\\hline
\multicolumn{1}{|c|}{Avg. vec.}&68.72&68.19&68.25&96.34&96.30&96.27&27.88&26.44&24.81\\\hline
\multicolumn{1}{|c|}{p-means}&\textit{72.20}&\textit{71.65}&\textbf{71.79}&96.69&96.67&96.65&33.77&33.41&33.26\\\hline
\multicolumn{1}{|c|}{ELMo}&71.20&\textbf{71.79}&71.36&94.54&91.32&91.32&\textit{42.35}&\textit{41.51}&\textit{41.54}\\\hline
\multicolumn{1}{|c|}{BERT}&70.89&70.79&70.88&95.52&95.39&95.39&39.92&39.38&39.35\\\hline
\multicolumn{1}{|c|}{EigenSent}&66.98&66.40&66.54&95.91&95.80&95.76&35.32&33.69&33.91\\\hline
\multicolumn{1}{|c|}{EigenSent$\oplus$Avg}&\textbf{72.24}&71.62&\textit{71.78}&\textbf{97.18}&\textbf{97.13}&\textbf{97.14}&\textbf{42.77}&\textbf{41.67}&\textbf{41.81}\\\hline\hline
\multicolumn{1}{|c|}{c[k]}&\textit{72.20}&71.58&71.73&\textit{96.98}&\textit{96.98}&\textit{96.94}&37.67&34.47&34.54\\\hline
\end{tabular}
\caption{Performance in text classification (20-NG, R-8) and sentiment (SST-5) tasks of various models as reported in \cite{kayal2019eigensent}, where DCT* refers to the implementation in \cite{kayal2019eigensent}. Our DCT embeddings are denoted as $c[k]$ in the bottom row. \textbf{Bold} indicates the best result, and \textit{italic} indicates second-best.}
\label{EigenSent}
\end{table*}
 
\subsection{Experimental setup}
For the word embeddings, we use pre-trained FastText embeddings of size 300 \cite{mikolov-etal-2018-advances} trained on Common-Crawl. We generate DCT sentence vectors by concatenating the first $K$ DCT coefficients, which we denote by $\mathbf{c[K]}$. We compare the performance against: vector averaging of the same word embeddings, denoted by \textbf{AVG}, and vector max pooling, denoted by \textbf{MAX}.\footnote{To compare with other sentence embedding models, refer to the results in \citet{conneau2018senteval} and \citet{conneau2018you}}

For all tasks, we trained multi-layer perceptron (MLP) classifiers following the setup in SentEval. We tuned the following hyper-parameters on the validation sets: number of hidden states (in [0, 50, 100, 200, 512]) and dropout rate (in [0, 0.1, 0.2]). Note that the case with 0 hidden states corresponds to a Logistic Regression classifier.
 
\subsection{Result \& Discussion}
We report the performance in probing tasks in Table \ref{ProbingBest}. In general, DCT yields better performance compared to averaging on all tasks, and larger $K$ often yields improved performance in syntactic and semantic tasks. For the surface information tasks, SentLen and Word content (WC), $c[0]$ significantly outperforms AVG. This is attributed to the non-linear scaling factor in DCT, where longer sentences are not discounted as much as in averaging. The performance decreased with increasing $K$ in $c[0:K]$, which reflects the trade-off between deep and surface linguistic properties, as discussed in \citet{conneau2018you}. %

While increasing $K$ has no positive effect on surface information tasks, syntactic and semantic tasks demonstrate performance gains with larger $K$. 
This trend is clearly observed in all syntactic tasks and three of the semantic tasks, where DCT performs well above AVG and the performance improves with increasing $K$. The only exception is SOMO, where increasing $K$ actually results in lower performance, although all DCT results are still higher than AVG. 

The correlation between the performance in probing tasks and the standard text classification tasks is discussed in \citet{conneau2018you}, where they show that most tasks are only positively correlated with a small subset of semantic or syntactic features, with the exception of TREC and some sentiment classification benchmarks. Furthermore, some tasks like SST and SICK-R are actually negatively correlated with the performance in some probing tasks like SubjNum, ObjNum, and BShift. This explains why simple averaging often outperforms more complex models in these tasks. Our results in Table \ref{downstream} are consistent with these observations, where we see improvements in most tasks, but the difference is not as significant as the probing tasks, except in TREC question classification where increasing $K$ leads to much better performance.  
As discussed in \citet{aldarmaki2018evaluation}, the ability to preserve word order leads to improved performance in TREC, which is exactly the advantage of using DCT instead of AVG. Note also that increasing $K$, while preserves more information, leads to increasing the number of model parameters, which in turn may negatively affect the generalization of the model by overfitting. In our experiments, $1 \le k \le 2$ yielded the best trade-off. 
\subsection{Comparison w. Related Methods}
Spectral analysis is frequently employed in signal processing to decompose a signal into separate frequency components, each revealing some information about the source signal, to enable analysis and compression. To the best of our knowledge, spectral methods have only been recently exploited to construct sentence embedding \cite{kayal2019eigensent}.\footnote{ independent from and in parallel with this work }.

\citeauthor{kayal2019eigensent} propose EigenSent that utilized Higher-Order Dynamic Mode Decomposition (HODMD) \cite{le2017higher} to construct sentence embedding. These embeddings summarize the dynamic properties of the sentence. In their work, they compare EigenSent with various sentence embedding models, including a different implementation of the Discrete Cosine Transform (DCT*). In contrast to our implementation described in section \ref{detailed}, DCT* is applied at the word level along the word embedding dimension.

For fair comparison, we use the same sentiment and text classification datasets, the SST-5, 20 newsgroups (20-NG) and Reuters-8 (R-8), as those used in \newcite{kayal2019eigensent}. We also evaluate using the same pre-trained word embedding, framework and approaches as described in their work. 
Table \ref{EigenSent} shows the best results for the various  models as reported in \newcite{kayal2019eigensent}, in addition to the best performance of our model denoted as $c[k]$.\footnote{The best results were achieved with k=3 for SST-5 and k=2 for 20-NG and R-8.} 

Note that the DCT-based model, DCT*, described in \newcite{kayal2019eigensent} performed relatively poorly in all tasks, while our model achieved close to state-of-the-art performance in both the 20-NG and R-8 tasks. Our model outperformed EignSent on all tasks and generally performed better than or on par with p-means, ELMo, BERT, and EigenSent$\oplus$Avg on both the 20-NG and R-8. On the other hand, both EigenSent$\oplus$Avg and ELMo performed better than all other models on SST-5.

\section{Conclusion}
We proposed using the Discrete Cosine Transform (DCT) as a mechanism to efficiently compress variable-length sentences into fixed-length vectors in a manner that preserves some of the structural characteristics of the original sentences. By applying DCT on each feature along the word embedding sequence, we efficiently encode the overall feature patterns as reflected in the low-order DCT coefficients. We showed that these DCT embeddings reflect average semantic features, as in vector averaging but with a more suitable normalization, in addition to syntactic features like word order. Experiments using the SentEval suite showed that DCT embeddings outperform the commonly-used vector averaging on most tasks, particularly tasks that correlate with sentence structure and word order. Without compromising practical efficiency relative to averaging, DCT provides a suitable mechanism to represent both the average of the features and their overall syntactic patterns. %
\bibliography{emnlp-ijcnlp-2019}
\bibliographystyle{acl_natbib}
\end{document}